\documentclass[conference]{IEEEtran}
\IEEEoverridecommandlockouts
\usepackage{cite}
\usepackage{amsmath,amssymb,amsfonts}
\usepackage{algorithmic}
\usepackage{graphicx}
\usepackage{textcomp}
\usepackage{xcolor}
\usepackage{booktabs}
\usepackage{url}
\def\BibTeX{{\rm B\kern-.05em{\sc i\kern-.025em b}\kern-.08em
    T\kern-.1667em\lower.7ex\hbox{E}\kern-.125emX}}
\begin{document}

\title{Dönüştürücü Tabanlı Doğal Dil İşleme Yöntemleri ile Reklam Metni Sınıflandırması\\
{\LARGE Ad Text Classification with Transformer-Based Natural Language Processing Methods}
}
\makeatletter
\newcommand{\linebreakand}{%
  \end{@IEEEauthorhalign}
  \hfill\mbox{}\par
  \mbox{}\hfill\begin{@IEEEauthorhalign}
}
\makeatother
\author{\IEEEauthorblockN{Umut Özdil}
\IEEEauthorblockA{\textit{Ar-Ge Departmanı} \\
\textit{AdresGezgini A.Ş.}\\
İzmir, Türkiye \\
umutozdil@adresgezgini.com}
\and
\IEEEauthorblockN{Büşra Arslan}
\IEEEauthorblockA{\textit{Ar-Ge Departmanı} \\
\textit{AdresGezgini A.Ş.}\\
İzmir, Türkiye \\
busraarslan@adresgezgini.com}
\and
\IEEEauthorblockN{Davut Emre Taşar}
\IEEEauthorblockA{\textit{Büyük Veri ve İleri Analitik Bölüm} \\
\textit{Garanti BBVA Teknoloji}\\
İstanbul, Türkiye \\
emretasa@garantibbva.com.tr}
\linebreakand 
\IEEEauthorblockN{Gökçe Polat}
\IEEEauthorblockA{\textit{Ar-Ge Departmanı} \\
\textit{AdresGezgini A.Ş.}\\
İzmir, Türkiye \\
gokcepolat@adresgezgini.com}
\and
\IEEEauthorblockN{Şükrü Ozan}
\IEEEauthorblockA{\textit{Ar-Ge Departmanı} \\
\textit{AdresGezgini A.Ş.}\\
İzmir, Türkiye \\
sukruozan@adresgezgini.com}
}
\maketitle
\renewcommand*\abstractname{Öz}
\begin{abstract}
Bu çalışmada, çevrimiçi reklam platformlarında oluşturulan reklam metinlerinin sektöre göre otomatik olarak sınıflandırılması için doğal dil işleme (NLP) tabanlı bir yöntem önerilmiştir. Eğitim veri setimiz 12 farklı sektöre ait yaklaşık 21.000 etiketli reklam metninden oluşmaktadır. Çalışmada son yıllarda doğal dil işleme literatüründe metin sınıflandırma gibi alanlarda sıkça kullanılan dönüştürücü (transformers) tabanlı bir dil modeli olan Transformatörlerden Çift Yönlü Kodlayıcı Gösterimleri (BERT) modeli kullanılmıştır. Türkçe diline yönelik olarak önceden eğitilmiş olarak seçilmiş bir BERT modelini kullanarak elde edilen sınıflandırma başarımları detaylı olarak gösterilmiştir.
\end{abstract}
\renewcommand*\IEEEkeywordsname{Anahtar Sözcükler}
\renewcommand*\abstractname{Abstract}
\begin{abstract}
In this study, a natural language processing-based (NLP-based) method is proposed for the sector-wise automatic classification of ad texts created on online advertising platforms. Our data set consists of approximately 21,000 labeled advertising texts from 12 different sectors. In the study, the Bidirectional Encoder Representations from Transformers (BERT) model, which is a transformer-based language model that is recently used in fields such as text classification in the natural language processing literature, was used. The classification efficiencies obtained using a  pre-trained BERT model for the Turkish language are shown in detail.
\end{abstract}
\renewcommand*\IEEEkeywordsname{Keywords}

\section{GİRİŞ} \label{sec:giris}
Yıllardır reklam sektörünün önde gelen medyaları olan TV ve gazete gibi geleneksel yayın organları günümüzde artık etkisini kaybetmeye başlamıştır. eMarketer'ın verilerine göre çevrimiçi reklamcılığının günümüzde en büyük reklam alanı olması beklenmektedir \cite{eMarkete2014}. İnternet ekonomisinin hızla büyümesi, reklam sektörünün dönüşümünde oldukça büyük bir etkiye sahiptir. 

Gelişen teknoloji ve internetin yaygın kullanımı dijital pazarlamayı, ürün ve hizmetlerini internet üzerinden tüketicilere sunmak isteyen işletmeler için önemli bir araç haline getirmekte, ilgilenmesi muhtemel potansiyel kişilere gösterilmesinin yanı sıra reklam performansının takibine ve analizine olanak sağlamaktadır\cite{Adsa2021}. Tüm bunlarla birlikte çevrimiçi reklamcılıkta, yatırım ve dönüşüm ilişkisinin incelendiği etkili reklam çalışmaları üzerine birçok araştırma yapılmaya başlanmıştır. 

Bu araştırmalar kapsamında insanların fikirlerini, tutumlarını, duygularını, varlıklara ve bireylere yönelik dosyalamalarını analiz etmek, web sitesi içerikleri, reklam metinleri, kullanıcı hareketleri, cihaz bilgisi, konum bilgisi veya demografik veriler gibi veriler üzerinden anlamlı sonuçlar çıkarmak için Makine Öğrenimi Algoritmaları ve Doğal Dil İşleme (NLP) teknikleri kullanılmaya başlanmıştır \cite{Abzetdin2016}.

Kaliteli bir reklam çalışmasının en önemli özelliği tüketiciye sunulan içeriğin alakalı ve etkili olmasıdır. Çevrimiçi reklamcılık platformlarında reklam kalitesi; reklam metinlerinin aramalarla ne kadar alakalı olduğu, tüketicilerin reklama tıklanma olasılığı ve reklam tıklandıktan sonra görüntülenen sayfada (landing page) yaşanan deneyim gibi birçok farklı faktöre bağlıdır. Yüksek reklam kalitesi daha iyi reklam konumu ve daha düşük maliyet ile sonuçlanmaktadır\cite{Adsb2021}. Bu noktada tüketici ihtiyaçlarına ve aramalarına cevap veren reklam metinlerinin hazırlanması, reklamverenin sunduğu ürünün/hizmetin potansiyel müşterilere etkili bir şekilde aktarılmasında önemli bir rol oynamaktadır. 

Çalışmamızda reklam kalitesini artırdığı bilinen reklam metinlerinin hangi sektör ile alakalı olduğunu tespit etmek amacıyla bir sınıflandırma yöntemi önerilmiştir. Çalışma gerçekleştirilirken, benzer sınıflandırma problemlerinde Word2vec, LSTM gibi modellerinden daha iyi sonuç verdiği \cite{Ozan2021} çalışmasında da irdelenmiş olan dönüştürücü (transformers) tabanlı bir dil modeli olan Transformatörlerden Çift Yönlü Kodlayıcı Gösterimleri (BERT) \cite{Devlin2018} kullanılmıştır.

\section{LİTERATÜR ANALİZİ} \label{sec:literatur}
Son yıllarda bilgisayarların hesaplama kapasitlerinin üstel bir şekilde artması ve yapay zeka ile derin öğrenme alanlarında hızlı gelişmelere paralel olarak, NLP alanında da birçok yeni mimari yapı geliştirilmiştir. Sinir ağlarının, verilerden özniteliklerini öğrenme konusunda yüksek kapasitelerinin olması sayesinde derin öğrenme, NLP alanının yanında birçok alanda araştırma konusu olmuştur. Ana sinir ağları, doğru bağlamı bulmak için sıklıkla kullanılan Uzun Kısa Süreli Bellek (LSTM) \cite{Hochreite1997}, dikkat mekanizması \cite{Chen2017} ve bellek ağlarıdır \cite{Rajpurkar2018}. Yinelemeli sinir ağları \cite{Dong2014}, kapalı sinir ağları  \cite{Xue2018} ve evrişimli sinir ağları  \cite{Huang2018} da NLP alanında kullanılan sinir ağlarındandır.

Nergiz vd., \cite{Nergiz2019} çalışmasında, Fasttext, Word2vec ve Doc2vec modellerinin Türkçe haber sitelerinde yer alan farklı kategorilere ait haber metinleri üzerindeki başarı oranlarını LSTM sinir ağı ile test ederek karşılaştırmış ve en başarılı sonucu Fasttext modelinin sağladığını sunmuştur.

Doğru vd. \cite{Dogru2021} çalışmasında, Türkçe ve ingilizce haber metinleri üzerinde Doc2vec kelime gömme yöntemi ile eğitim modeli oluşturulmuş, derin öğrenme sınıflandırma yöntemi CNN ve makine öğrenmesi sınıflandırma yöntemleri Gauss Naive Bayes (GNB), Random Forest (RF), Naive Bayes (NB) ve Support Vector Machine (SVM) ile modelin doğruluk oranlarını karşılaştırarak en yüksek sonucun CNN ile yapılan sınıflandırmada elde edildiği gösterilmiştir.

González-Carvajal ve Garrido-Merchán,\cite{GonzalezCarvajal2020} çalışmasında, sosyal medyada paylaşılan iletiler, film-dizi eleştirileri, haber içerikleri gibi birçok alan üzerinde BERT modeli ve geleneksel doğal dil işleme yaklaşımlarının karşılaştırmalı analizlerini gerçekleştirmiş ve sonuçlarını sunmuştur. Sonuçlar BERT’in geleneksel NLP yaklaşımlarından farklı alanlarda daha iyi sonuçlar verdiği ampirik değerler ile kanıtlanmıştır. 

Zhengjie vd.\cite{Zhengjie2019} çalışmasında duygu analizine bağlı sınıflandırma yapmak için BERT modelinin hedefe bağlı üç varyasyonu kullanılmıştır. Üç veri kümesi üzerinde yapılan deneyler sonucunda modelin cümle düzeyinde duyarlılık sınıflandırması dahil olmak üzere birçok NLP probleminde mevcut yöntemlere göre daha iyi sonuçlar verdiği gösterilmiştir.

\section{TASARIM VE YÖNTEM} \label{sec:tasarımveyöntem}

\subsection{Veri Seti}\label{sec:veriseti}
Kullanılan veri seti, 12 farklı iş sektörüne ait yaklaşık 21.000 reklam metninden oluşmaktadır. Ön işleme aşamasında veri set öncelikle tekrar eden verilerden arındırılmıştır. Noktalama işaretleri ve semboller çıkarılmıştır.
Veri setinde kullanılan 12 temel kategori için rastgele seçilmiş üç reklam metni Tablo \ref{tablo:ornekveri}'de görülmektedir.  Bunun yanında Şekil.\ref{img:sekil1} ve Şekil. \ref{img:sekil2} den de anlaşılabileceği gibi her bir kategorinin farklı sayıda reklam metnine sahip olması ve reklam metinlerinin farklı sayıda kelimeye sahip olması kullanılan veri setindeki dengesizliği göstermektedir. Bu dengesiz veri seti ile eğitilmiş BERT modelinden alınan kategorik olasılıksal sınıflandırma yönteminin çıktıları analiz edilerek üstün bir başarı gösterdiği gözlemlenmiştir. 
\begin{figure}
	\centering\includegraphics[width=0.48\textwidth]{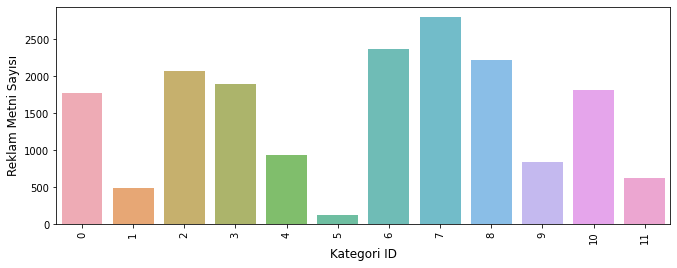}
	\caption{Her bir kategoride bulunan reklam metin sayısı}
	\label{img:sekil1}
\end{figure}

\begin{figure}
	\centering\includegraphics[width=0.48\textwidth]{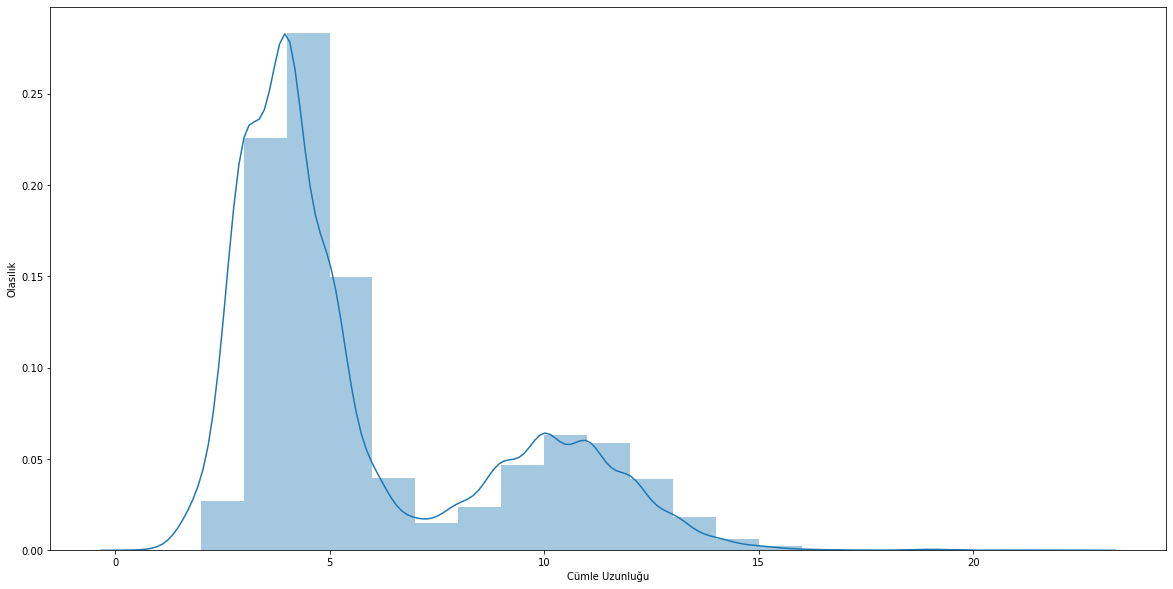}
	\caption{Tüm reklam metinlerindeki kelime sayısı dağılımı}
	\label{img:sekil2}
\end{figure}

\begin{table}[h!]
  \centering
  \caption{\textsc{Rastgele Seçilmiş Veri Seti Örneği}}
  \label{tablo:ornekveri}
  \begin{tabular}{|l|l|l|}
    \hline
    \textbf{\scriptsize ID} & \textbf{ \scriptsize Kategori} & \textbf{\scriptsize Reklam Metni}\\
   \hline
                  &                              & \scriptsize Çiçek ve Çikolata Hediyeleri\\
    \scriptsize 0 &  \scriptsize Çiçek Siparişi  & \scriptsize Özel Günlerinize, Özel Çiçekler\\
                  &                              & \scriptsize Eşsiz Çiçek Tasarımları\\
    \hline
                  &                                     & \scriptsize Hayvan Dostunuz için En İyisi\\
    \scriptsize 1 &  \scriptsize Evcil Hayvan Ürünleri  & \scriptsize Köpek Maması ve Vitaminleri\\
                  &                                     & \scriptsize Türkiye'nin Rakipsiz Petshop'u\\
    \hline
                  &                                          & \scriptsize Özel Taşlı Bileklikler\\
    \scriptsize 2 &  \scriptsize Mücevher, Takı \& Aksesuar  & \scriptsize Size En Yakışan Saat Burada! \\
                  &                                          & \scriptsize En Ucuz \& En Yeni Kolye Modelleri.  \\
    \hline
                  &                              & \scriptsize Nakliye ve Depolamada Öncü Firma\\
    \scriptsize 3 & \scriptsize Nakliyat, Kargo  & \scriptsize Güvenilir ve Hızlı Kargo \\
                  &                              & \scriptsize Paketleyerek Profesyonel Taşıma.  \\ 
    \hline
                  &                                         & \scriptsize Son Teknoloji Hidrolik Yağlar\\
    \scriptsize 4 & \scriptsize Oto Aksesuar \& Yedek Parça & \scriptsize Araç Koltuk Temizliği \\
                  &                                         & \scriptsize Aradığınız Tüm Jant Çeşitleri  \\  
    \hline
                  &                                             & \scriptsize Oyunun Heyecanına Katılın\\
    \scriptsize 5 & \scriptsize Oyunlar,                        & \scriptsize En iyi Korku Evi\\
                  & \scriptsize Oyun Konsolları \&  Ekipmanları & \scriptsize Oyunlarda Uygun Fiyatlar!  \\  
    \hline
                  &                                      & \scriptsize İlaçsız Terapi Yöntemi\\
    \scriptsize 6 & \scriptsize Psikolojik Danışmanlık   & \scriptsize Öfke Kontrolü Yöntem \& Tedavi\\
                  &                                      & \scriptsize Kişiye Özel Klinik Terapi Yöntemleri  \\  
    \hline
                  &                                      & \scriptsize Profesyonel Yıkama Hizmetleri\\
    \scriptsize 7 & \scriptsize Temizlik \& Halı Yıkama  & \scriptsize Temizlik Malzemeleri ve Aparatlar.\\
                  &                                      & \scriptsize Uygun Fiyatlarla Mükemmel Temizlik! \\  
    \hline
                  &                              & \scriptsize Yurtiçi \& Yurtdışı Tur Fırsatı\\
    \scriptsize 8 & \scriptsize Tur Acenteleri   & \scriptsize Avantajlı Otel \& Tatil Fırsatı\\
                  &                              & \scriptsize Aradığınız Konaklama Fırsatları \\  
    \hline  
                  &                             & \scriptsize Hızlı \& Kolay Vize İşlemleri\\
    \scriptsize 9 & \scriptsize Vize İşlemleri  & \scriptsize Evraksız Rusya Vizesi\\
                  &                             & \scriptsize Vize İşlemlerinin Doğru Adresi \\  
    \hline  
                   &                                  & \scriptsize Müfredata Uygun Eğlenerek Öğrenme\\
    \scriptsize 10 & \scriptsize Yabancı Dil Eğitimi  & \scriptsize Erken Yaşta İngilizce Dil Eğitimi\\
                   &                                  & \scriptsize Konuşma Odaklı Drama ile İngilizce \\  
    \hline  
                   &                     & \scriptsize Ücretsiz wireless \& Çalışma Odaları\\
    \scriptsize 11 & \scriptsize Yurtlar & \scriptsize Evinizdeki Konfor için Bize Ulaşın!\\
                   &                     & \scriptsize 7/24 Sıcak Su \& Güvenlik Hizmeti.\\ 
    \hline
  \end{tabular}
\end{table}

\subsection{Yöntem} \label{subsec:yöntem} 

\subsubsection{BERT} \label{subsubsec:bert}

Transformatörlerden Çift Yönlü Kodlayıcı Gösterimleri (BERT), etiketlenmemiş metinden çift yönlü gösterimleri önceden eğitmek ve sonrasında farklı NLP görevleri için etiketli metin kullanılarak ince ayar yapmak için tasarlanmış \cite{Devlin2018} ve Google tarafından 2018 yılında sunulmuş bir dönüştürücü modelidir. Dönüştürücüler, kuyruk yapısıyla oluşturulmuş öz dikkat mekanizması ile çalışan bir yapıya sahiptir Şekil. \ref{img:sekil3}. BERT modeli, bir sorguyu ve bir dizi anahtar-değer çiftini bir çıktıya eşler. Burada sorgu, anahtarlar, değerler ve çıktının kendi arasındaki korelasyonu ifade edebilecek vektörler oluşmaktadır. Çıktı, değerlerin ağırlıklı toplamı ve bir değere atanan ağırlık, sorguya karşılık gelen anahtarla uyumluluk oranı ile hesaplanır \cite{Vaswani2017}. BERT modeli, bir metni hem sağdan sola hem de soldan sağa işlemektedir, bu sayede metin içerisindeki öğeler arasındaki ilişkileri öğrenebilmektedir.  Eğitim aşamasında, MLM (Masked Language Modeling) ve NSP (Next Sentence Prediction) teknikleri kullanılmaktadır. MLM tekniğinde, maskelenen kelimeler, açık (maskelenmeyen) kelimeler kullanılarak tahmin edilmeye çalışılmaktadır. Bu teknik ile cümle içindeki kelimeler üzerinden inceleme ve tahminleme yapılmaktadır. NSP tekniğinde, cümlelerin birbirleri ile ilişkisi incelenmektedir. Bir cümlenin kendisinden sonra gelen cümle ile ilişkisi incelenmektedir. BERT modeli ile kurulan yapılarda önceden eğitilmiş bir model gereksinimi vardır. Bu sebeple çalışmamızda Loodos takımı tarafından Türkçe dili için önceden MLM tekniği ile eğitilmiş bir BERT modeli olan BERT-BASE-TURKISH-UNCASED \cite{Loodos2020} modelini tercih ettik. Kullanılan modelin ön eğitim parametreleri Tablo \ref{tablo:onparametre}’de verilmiştir. Bu model üzerinde ince ayar işlemi yapılarak kullanıma hazır hale getirilmiştir.

\begin{figure}
	\centering
		\includegraphics[width=0.50\textwidth]{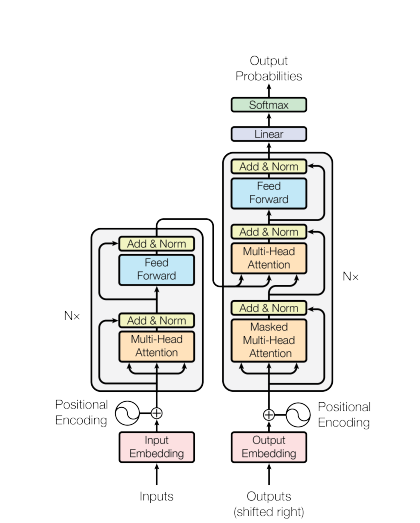}
	\caption{Dönüştürücü Model Mimarisi \cite{Vaswani2017}}
	\label{img:sekil3}
\end{figure}

\begin{table}[h!]
  \centering
  \caption{\textsc{Kullanılan Ön Eğitimli BERT Modeinin Parametreleri}}
  \label{tablo:onparametre}
  \begin{tabular}{|l|l|l|l|l|l|l|}
  \hline  
	\scriptsize Model& \begin{tabular}[c]{@{}c@{}} \scriptsize Gizli\\ \scriptsize Katman\\ \scriptsize Boyutu\end{tabular} & \begin{tabular}[c]{@{}c@{}} \scriptsize Maks.\\ \scriptsize Sekans\\ \scriptsize Uzunluğu\end{tabular} &\begin{tabular}[c]{@{}c@{}} \scriptsize Dikkat \\\scriptsize Ana\\\scriptsize Başlığı\\\scriptsize Sayısı\end{tabular} & \begin{tabular}[c]{@{}c@{}} \scriptsize Gizli\\\scriptsize Katman \\\scriptsize Sayısı\end{tabular} & \scriptsize Mimari                                                     & \begin{tabular}[c]{@{}c@{}}Sözlük\\ Boyutu\end{tabular} \\
	\hline
	\begin{tabular}[c]{@{}c@{}}\scriptsize (Loodos\\ 2020)\end{tabular} & \scriptsize 768                                                             & \scriptsize 512                                                                  &  \scriptsize12                                                                    & \scriptsize 12                                                               & \begin{tabular}[c]{@{}c@{}} \scriptsize Bert\\ \scriptsize For\\ \scriptsize Masked\\ \scriptsize LM\end{tabular} & \scriptsize 32000\\  
   \hline
   \multicolumn{7}{l}{\tiny *Gizli Katman Boyutu:~Kodlayıcı katmanlarının ve havuzlama katmanlarının boyutları} \\
   \multicolumn{7}{l}{\tiny Maksimum Sekans Uzunluğu: Modelin kullanabileceği maksimum sekans uzunluğu} \\
   \multicolumn{7}{l}{\tiny Dikkat Ana Başlığı Sayısı: Dönüştürücü kodlayıcısı içerisindeki her dikkat katmanı için dikkat ana başlığı sayısı.} \\
   \multicolumn{7}{l}{\tiny Gizli katman Sayısı: Dönüştürücü kodlayıcısı içerisindeki gizli katmanların sayısı.}
   \end{tabular}
\end{table}
\subsubsection{F1 Skoru ve Başarım Kriterleri} \label{subsubsec:fskor}

Eğitilen modelin başarısı: doğruluk değeri, F1 skoru ve hata dizeyi (confusion matrix)  ile ölçülmüştür.  Başarı ölçümleri: True Positive (TP), False Positive (FP), True Negative (TN) ve False Negative (FN) değerleri ile ölçülmektedir. TP, modelin tahmini ve gerçek değerin her ikisininde olumlu sonuç vermesi TN, modelin tahmini ve gerçek değerin her ikisininde olumsuz sonuç vermesi, FP model tahmini olumlu iken gerçek değerin olumsuz sonuç vermesi ve FN  ise modelin tahmini olumsuz iken gerçek değerin olumlu sonuç vermesi şeklinde açıklanabilir. Bu durumda  TP ve TN doğru sonuç, FP ve FN ise yanlış sonuç kabul edilir. 

Doğruluk değeri hesaplanırken, modelin doğru tahmin ettiği TP ve TN değerlerinin, tahmin edilen tüm TP, TN, FP, FN değerlerine oranı ile hesaplanmaktadır (\ref{eq:accuracy}). Ancak çok sınıflı problemlerde bu kriter ile sınıf bazlı doğruluk değeri alınamamaktadır. Bu sıkıntıyı ortadan kaldırmak amacıyla F1 skoru adı verilen kesinlik-doğruluk değerlerine dayanan bir başarı kriteri kullanılmaktadır.     

\begin{equation} \label{eq:accuracy}
    Do\Breve{g}ruluk =  \frac{TP+TN}{TP+TN+FP+FN}
\end{equation}

Kesinlik (precision) değeri, modelin tahmin ettiği TP değerlerin sayısının, modelin ürettiği tüm olumlu sonuçlar olan TP ve FP değerlerinin sayısına oranıdır (\ref{eq:precision}). Duyarlılık değeri ise modelin tahmin ettiği TP değerlerinin sayısının, modelin üretmesi gereken tüm olumlu sonuçlar olan TP ve FN sayılarına oranı ile bulunabilir (\ref{eq:recall}). F1 skoru ise kesinlik ve duyarlılık değerlerinin harmonik ortalaması olarak tanımlanabilir (\ref{eq:f1score}). 

\begin{equation} \label{eq:precision}
    Kesinlik =  \frac{TP}{TP+FP}
\end{equation}

\begin{equation} \label{eq:recall}
    Duyarl\i l \i k = \frac{TP}{TP+FN}
\end{equation}

\begin{equation} \label{eq:f1score}
    F1 Skoru = 2\left (\frac{Kesinlik\times Duyarl\i l \i k}{Kesinlik+ Duyarl\i l \i k}\right)
\end{equation}

Çok sınıflı verilerin dengesiz dağıldığı çalışmalar, modelin başarımını doğru bir şekilde değerlendirmek için bir sorun teşkil etmektedir. Bu durumu ortadan kaldırmak adına F1 skorunun sınıf bazlı veri dağılımına göre ağırlıklı ortalamasının alınarak hesaplanabilir.

\section{BULGULAR VE TARTIŞMA}

Yapılan analizler sonucunda, çalışmada kullandığımız Türkçe dili için ön eğitime tabi tutulmuş BERT modeli, veri setimizle ince ayar işlemine tabi tutulmuştur. İnce ayar işlemi BERT kütüphanesindeki kategorik sınıflandırma yöntemi ile gerçekleştirilmiştir. Bu yöntemde veri seti önceden belirlenen kategorilere ayrılarak, veri setinin \%70’i eğitim veri seti,\%30’u ise modelin test edilmesi için test eğitim seti olarak kullanılmıştır. Eğitim-test ayrıştırması kategorik olarak yapıldığı için her kategoriden eşit oranda veri ile eğitim ve test gerçekleştirilmiştir. Daha sonra 14349 eğitim verisi ve 3588 test verisi ile model 10 iterasyonda eğitildikten sonra, modelin 3. iterasyonundan sonra, doğruluk kesinlik, duyarlılık ve F-1 değerlerinde bir artış olmamasından ötürü 3. iterasyonun sonuçları üzerinden analizler gerçekleştirilmiştir. Burada farklı eğitim - test seti bölünme oranları karşılaştırılmış olsa da literatürde yapılmış benzer çalışmalarda kullanılan \%70 - \%30 oranı ile uyumlu kalınarak sonuçların üretilmesi hedeflenmiştir. Test veri setindeki kategorilerin kesinlik değerleri arasındaki en düşük değer 25 adet test verisi içeren 5 numaralı kategori için gerçekleşmiştir. Duyarlılık analizinde yine en düşük değer 1 numaralı kategori için 97 adet test verisi olan sınıfta bulunmuştur. F-1 skoru için ise yine 5 numaralı kategoride minimum değerle karşılaşılmıştır. Modelin en düşük başarım gösteren kategorilerinin eğitim ve test veri seti içerisinde en düşük orana sahip olan verilerden oluştuğu görülmektedir. Buradan yola çıkarak benzer çalışmalar için ideal olan eğitim ve test verisinin kategori bazında 1000’den üzeri olduğu durumlarda \%90 üzerinde bir doğruluğa ulaşabileceği görülmektedir. Tablo \ref{tablo:yontemler}’te 3. iterasyona ait sınıflandırma raporu, Şekil \ref{img:resim4}’te ise ilk 3 iterasyon için olan hata dizeyleri görülmektedir. Burada da görüldüğü gibi model ilk iterasyondan itibaren yüksek bir eğitim doğruluğuna ulaşmakta, ancak, 3. iterasyondan sonra yüksek bir test doğruluğu kazanacak düzeye erişebilmektedir. 

\begin{table}[h!]
\centering
  \caption{\textsc{3.İterasyon Test Raporu}}
  \label{tablo:yontemler}
  \begin{tabular}{*5c}
    \toprule
    \textbf{ID} & \textbf{Kesinlik} & \textbf{Duyarlılık} & \textbf{F1 Skoru} & \textbf{Reklam Metinleri}\\
    \midrule
    0 & 0.90 & 0.95 & 0.92 & 353\\
    1 & 0.99 & 0.77 & 0.87 & 97\\
    2 & 0.92 & 0.92 & 0.92 & 414\\
    3 & 0.93 & 0.87 & 0.90 & 378\\
    4 & 0.88 & 0.87 & 0.87 & 187\\
    5 & 0.78 & 0.84 & 0.81 & 25\\
    6 & 0.95 & 0.94 & 0.90 & 474\\
    7 & 0.87 & 0.93 & 0.91 & 560\\
    8 & 0.92 & 0.90 & 0.94 & 444\\
    9 & 0.91 & 0.98 & 0.94 & 168\\
    10 & 0.94 & 0.93 & 0.94 & 363\\
    11 & 0.87 & 0.82 & 0.84 & 125\\
    \midrule
    Doğruluk & & & \textbf{0.91} & \textbf{3588} \\
    Ağ. Ort & \textbf{0.91}& \textbf{0.91} & \textbf{0.91} & \textbf{3588}\\
    \bottomrule

  \end{tabular}
\end{table}

\begin{figure*}[!b]
	\centering
		\includegraphics[width=0.75\textwidth]{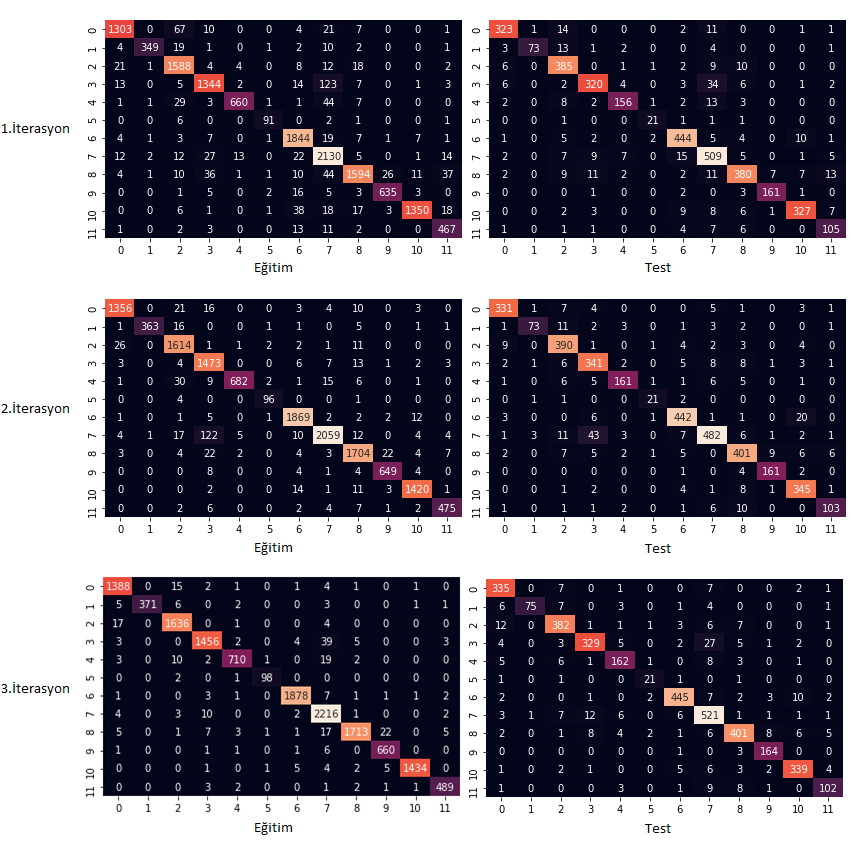}
	\caption{Eğitim ve Test Verilerine ait Hata Dizeyleri.}
	\label{img:resim4}
\end{figure*}

\section{SONUÇ, ÖNERİLER VE KISITLAR}
Çevrimiçi reklam platformlarında oluşturulan reklam metinlerinin sektöre göre otomatik olarak sınıflandırılması için gerçekleştirdiğimiz bu doğal dil işleme uygulaması, veri setimizdeki 12 farklı sektöre ait yaklaşık 21.000 etiketli reklam metnini \%90’ın üzerinde bir doğruluk ile ait olduğu kategoriye göre sınıflandırabilmiştir. Bu yöntemi gerçekleştirirken günümüzde doğal dil işleme alanında sıkça kullanılan dönüştürücü (transformers) tabanlı bir dil modeli olan Transformatörlerden Çift Yönlü Kodlayıcı Gösterimleri (BERT) modeli veri setimiz ile ince ayar yapılarak kullanılmıştır. Gerçekleştirdiğimiz bu sınıflandırma kapalı bir alan içerisinde ve önceden belirlenmiş kategoriler için gerçekleştirilmiştir. Yöntem, çevrimiçi reklam ağlarında kullanılacak olan metin tabanlı reklamların, reklamverenin sektörü ile uyumluluğunun test edilebilmesi için kullanılabilecektir. Bu sayede çevrimiçi reklam kalitesinin belirlenmesinde önemli bir bileşen olan reklam metninin sektör ile uygunluğu durumu otomatik bir şekilde test edilebilecektir.

\bibliographystyle{IEEEbib}
\bibliography{referanslar}

\end{document}